\documentclass[letterpaper]{article} 
\usepackage{aaai24}  
\usepackage{times}  
\usepackage{helvet}  
\usepackage{courier}  
\usepackage[hyphens]{url}  
\usepackage{graphicx} 
\usepackage{natbib}  
\usepackage{caption} 
\usepackage{algorithm}
\usepackage{algorithmic}

\usepackage{amsmath,amsthm,amssymb,mdframed}
\usepackage{newfloat}
\usepackage{listings}
\usepackage{booktabs}
\usepackage{makecell}
\usepackage{amsfonts} 
\usepackage{dsfont}
\usepackage{multirow}

\usepackage{newfloat}
\usepackage{listings}
\DeclareCaptionStyle{ruled}{labelfont=normalfont,labelsep=colon,strut=off} 
\floatstyle{ruled}
\newfloat{listing}{tb}{lst}{}
\floatname{listing}{Listing}
\title{Compositional Generalization for Multi-label Text Classification: A Data-Augmentation Approach}
\author{
    Yuyang Chai\textsuperscript{\rm 1,}\equalcontrib,
    Zhuang Li\textsuperscript{\rm 2,}\equalcontrib,
    Jiahui Liu\textsuperscript{\rm 1},
    Lei Chen\textsuperscript{\rm 1},
    Fei Li\textsuperscript{\rm 1},
    Donghong Ji\textsuperscript{\rm 1},
    Chong Teng\textsuperscript{\rm 1,}\thanks{Corresponding author.}\\
}
\affiliations{
    \textsuperscript{\rm 1}Key Laboratory of Aerospace Information Security and Trusted Computing, Ministry of Education, \\
    School of Cyber Science and Engineering, Wuhan University\\
    \textsuperscript{\rm 2}Faculty of Information Technology, Monash University\\

    \{yychai, liujiahui, chhenl, lifei\_csnlp, dhji, tengchong\}@whu.edu.cn, zhuang.li1@monash.edu
}

\usepackage{bibentry}

\usepackage{amsthm,amsmath,amsfonts,bm,xspace}
\usepackage{color}

\def\eqref#1{(\ref{#1})}

\def\1{\bm{1}}

\def\rmI{{\mathbf{I}}}

\def\rmM{{\mathbf{M}}}

\def\rmV{{\mathbf{V}}}
\def\rmW{{\mathbf{W}}}

\def\vtheta{{\bm{\theta}}}

\def\vh{{\bm{h}}}

\def\vk{{\bm{k}}}
\def\vl{{\bm{l}}}

\def\vv{{\bm{v}}}

\def\vx{{\bm{x}}}
\def\vy{{\bm{y}}}
\def\vz{{\bm{z}}}

\DeclareMathAlphabet{\mathsfit}{\encodingdefault}{\sfdefault}{m}{sl}
\SetMathAlphabet{\mathsfit}{bold}{\encodingdefault}{\sfdefault}{bx}{n}

\def\gL{{\mathcal{L}}}

\newcommand{\KL}{\mathbb{D}_{\mathrm{KL}}}

\newcommand{\vTheta}{\bm{\theta}}
\newcommand{\vPhi}{\bm{\phi}}

\newcommand{\vMu}{\bm{\mu}}
\newcommand{\vSigma}{\bm{\sigma}}

\DeclareMathOperator*{\argmax}{arg\,max}

\begin{document}
\maketitle

\begin{abstract}
Despite significant advancements in multi-label text classification, the ability of existing models to generalize to novel and seldom-encountered complex concepts, which are compositions of elementary ones, remains underexplored. This research addresses this gap. By creating unique data splits across three benchmarks, we assess the compositional generalization ability of existing multi-label text classification models. Our results show that these models often fail to generalize to compositional concepts encountered infrequently during training, leading to inferior performance on tests with these new combinations. To address this, we introduce a data augmentation method that leverages two innovative text generation models designed to enhance the classification models' capacity for compositional generalization. Our experiments show that this data augmentation approach significantly improves the compositional generalization capabilities of classification models on our benchmarks, with both generation models surpassing other text generation baselines\footnote{Codes available at https://github.com/yychai74/LD-VAE}.
\end{abstract}
\section{Introduction}
Multi-label text classification (MLTC) involves identifying the labels associated with the input text. This task has broad applications in natural language processing (NLP), including sentiment analysis in tweets~\cite{semeval}, subject identification in interdisciplinary academic articles~\cite{SGM}, and movie genre classification based on movie reviews~\cite{IMDB}. Although there has been significant progress in improving classifier performance across various MLTC benchmarks, whether existing MLTC models can generalize compositionally has received limited attention in prior MLTC studies.


\begin{figure}[!ht]
    \centering
    \includegraphics[width=0.75\linewidth]{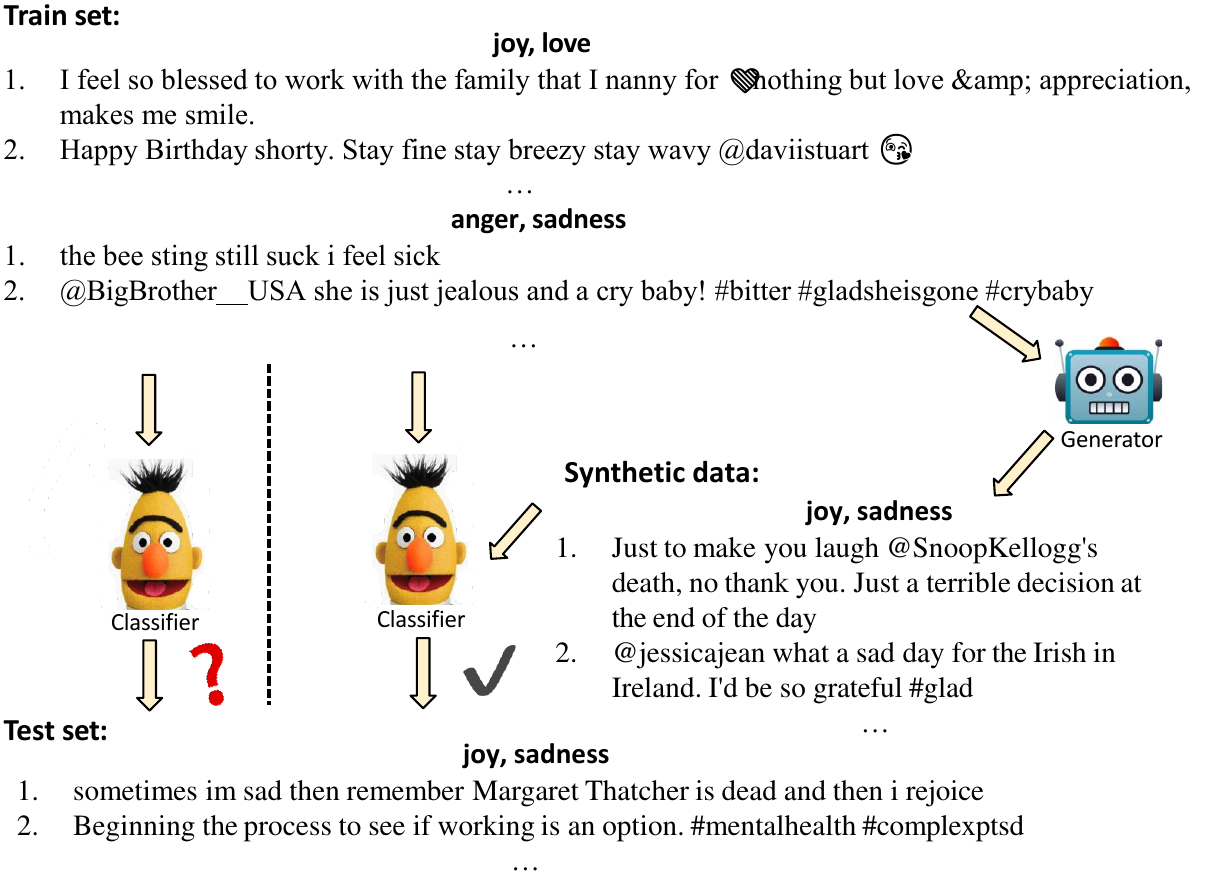}
    \caption{Illustration of the CG challenge in MLTC and an overview of our proposed data augmentation solution.
    }
    \label{intro}
    
\end{figure}

Compositional generalization (CG) is a fundamental ability inherent to human intelligence, enabling the recognition of novel and infrequently occurring high-level concepts that are compositions of more atomic elements~\cite{chomsky2014aspects}. For example, once a person learns to understand the emotions of joy and sadness in simple phrases like `I am sad' and `I rejoice', respectively, he can effortlessly recognize a complex emotion in a tweet such as `sometimes I am sad, then remember Margaret is dead, and then I rejoice'. This tweet conveys a nuanced composite of two emotions occurring simultaneously. In contrast to humans, our initial research indicates that current MLTC models struggle to identify these nuanced compositions if they infrequently occur in the training set. The T5-based~\cite{raffel2020exploring} MLTC model~\cite{clp}, for example, only achieved less than 2\% accuracy on the SemEval test set~\cite{semeval} for previously unseen emotional compositions, despite ample training data for each elementary emotion. Conversely, when the training data contains abundant emotional compositions as those found in the test set, its accuracy exceeded 28\%. This discrepancy underscores the urgent need for MLTC models capable of generalizing to novel compositions, making them more effective in a world that continuously presents new composite knowledge.


This study offers the first in-depth exploration of the CG challenges that impact MLTC. We utilize three MLTC benchmarks that tackle three tasks: emotion classification, subject identification of abstracts, and genre classification of movie reviews. Traditional MLTC benchmarks typically employ random splits, where all the compositions of individual labels are prevalent across both training and test sets. This methodology hinders a rigorous evaluation of the models' capacity to generalize compositionally. Inspired by the inherent human ability to recognize unfamiliar composite concepts with minimal exposure, as well as informed by previous research on CG~\cite{keysers2019measuring,finegan2018improving}, we propose a distinct data split to evaluate the CG capabilities of MLTC models. All elementary labels are abundant in the training set in this split, but instances with novel label compositions in the test set are seldom found in the training data. To enhance this evaluation, we introduce two novel metrics that evaluate model performance in terms of compositional rather than individual label predictions.



Compositional data augmentation is a widely-adopted strategy for enhancing the CG ability of machine learning models in fields such as semantic parsing~\cite{qiu2022improving,yang2022subs,andreas2020good} and few-shot single-label classification~\cite{li-etal-2022-variational-autoencoder}. This strategy augments training sets with instances of seldom-occurring compositions. Inspired by the methodology, we introduce an innovative data augmentation technique specifically designed for MLTC. Our approach consists of three key components: i) a model that learns the distribution of target label compositions using limited examples, ii) a conditional generation model that synthesizes new text instances conditioned on novel label compositions drawn from the learned distribution, and iii) a filter to discard invalid examples.

A fundamental challenge is devising a generation model that can identify and systematically combine phrases tied to individual labels, thereby forming coherent text that reflects the given label composition. A significant hurdle is the entangled representations of individual semantic factors in neural sequence models, hindering the alignment learning of labels with corresponding text fragments. Therefore, We propose two innovative representation disentanglement solutions for conditional generation models. The first, Label-specific Prefix-Tuning (LS-PT), uses label-specific prefix vectors~\cite{li2021prefix} as disentangled label representations.
The second, Label Disentangled Variational Autoencoder (LD-VAE), employs disentanglement learning and variational autoencoders~\cite{kingma2013auto} to extract disentangled label representations. Conditioned on the disentangled label representations, the conditional generation models can yield high-quality instances.

Overall, our contributions are three-fold:


\begin{itemize}
 \item We are the first to explore the critical issue of CG on three MLTC benchmarks. By introducing a unique evaluation data split and two novel evaluation metrics, we can measure the CG abilities of existing models. Our analysis reveals existing MLTC models lack CG capability.

 \item  We propose a novel data augmentation strategy to augment instances with rare label compositions. Our empirical studies demonstrate that this approach with various generation models dramatically boosts the CG capability of MLTC models across all evaluated metrics.

 \item  We design two generation models central to our data augmentation approach. These models focus on disentangling and composing individual label representations to generate instances associated with novel label compositions. Experiments show that both models surpass other generation baselines regarding CG evaluation metrics.
\end{itemize}

\section{Compositional Multi-label Text Classification}

\paragraph{Problem Setting.}
In MLTC, the aim is to determine a function $\pi_{\theta}: \mathcal{X} \rightarrow \mathcal{Y}$, that maps a text sequence $\vx \in \mathcal{X}$ to its corresponding label set $\vy \in \mathcal{Y}$. Here, $\vx = \{x_0,...,x_n\}$ represents a sequence of $n$ tokens, while $\vy = \{y_0,...,y_m\}$ denotes a composition of $m$ unordered labels in a label set.

We assume that training set samples $\langle \vx, \vy \rangle \in \mathcal{D}_{train}$ are drawn from a source distribution $P_s(\vx, \vy)$, while test set samples $\mathcal{D}_{test}$ are sourced from a target distribution $P_t(\vx, \vy)$. In the conventional training and evaluation setting, where the dataset is split randomly, both distributions align, i.e., $P_s(\vx, \vy) = P_t(\vx, \vy)$. However, in our compositional data split, which is in line with the CG studies for other tasks~\citep{qiu2022improving,yang2022subs,andreas2020good}, these distributions diverge. We assume the conditional distribution remains the same, i.e., $P_s(\vx|\vy) = P_t(\vx|\vy)$, while the label composition distribution varies: $P_s(\vy) \neq P_t(\vy)$. In addition, in the CG setup, the atomic individual labels $y \in Y$ are \textit{shared and abundant} across both training and test datasets. Additionally, an optional support set $\mathcal{D}_{s}$ complements the training set, comprised of limited examples drawn from $P_t(\vx, \vy)$. This aids in the few-shot learning of novel compositions, as seen in the setups of the CG works~\cite{lee2019one,li2021few}. 

Concretely, the training, support, and test sets are constructed as follows. Let the complete MLTC dataset be denoted as $\mathcal{D}_{ori} = \mathcal{D}_{train} \cup \mathcal{D}'_{test}$. We partition $\mathcal{D}_{ori}$ into training and preliminary test sets based on label compositions. Specifically, $\mathcal{D}_{train} = \mathcal{X}_{train} \times \mathcal{Y}_{train}$ and $\mathcal{D}'_{test} = \mathcal{X}'_{test} \times \mathcal{Y}'_{test}$. We ensure that the training and preliminary test sets do not share any label compositions: $\mathcal{Y}_{train} \cap \mathcal{Y}'_{test} = \emptyset$. The preliminary test set contains $M$ unique label compositions, where $|\mathcal{Y}'_{test}|=M$. 
We then randomly sample a subset of $\mathcal{D}'_{test}$ containing $N_{s}$ examples to form the support set, denoted as $\mathcal{D}_s$. The remaining examples in $\mathcal{D}'_{test}$ constitute the actual test set, $\mathcal{D}_{test}=\mathcal{D}'_{test} \setminus \mathcal{D}_s$. Given that we employ a data augmentation module, the models within this module are trained on the union of the training and support sets, $\mathcal{D}_{cg} = \mathcal{D}_s \cup \mathcal{D}_{train}$. This data augmentation process augments $\mathcal{D}_s$ to yield a synthetic data set, denoted as $\mathcal{D}_{aug}$. Our multi-label classifier is trained on the combined training, support, and augmented sets, $\mathcal{D}_{mltc} = \mathcal{D}_s \cup \mathcal{D}_{train}\cup\mathcal{D}_{aug}$, and it is evaluated on $\mathcal{D}_{test}$.

\paragraph{Evaluation Metrics.}
To assess MLTC performance, one of the prevailing metrics is the \textbf{Jacc}ard score, as highlighted in~\citet{SGM,clp}. The Jaccard score computes the ratio of correctly identified labels to the union of the two label sets for each instance. However, this metric might inadequately capture the CG capability since it measures performance based on individual labels. Hence, even if a predicted label composition is erroneous, a high Jaccard score can still be attained. To capture composition-level performance, we use \textbf{Acc}uracy metric as in ~\citet{10.1007/978-3-030-72610-2_14}. This metric verifies if the predicted label set exactly matches the ground-truth set.
While the Accuracy provides valuable insights at the composition level, we anticipate a more nuanced analysis. Therefore, we introduce two supplementary metrics, \textbf{Corr}ectness and \textbf{Comp}leteness, which evaluate performance at the compositional level but with a degree of flexibility. The Correctness metric evaluates if every predicted label exists in the ground truth, whereas Completeness checks if every ground-truth label has been forecasted by the model:
\begin{align}
\text{Corr}(\vy_{p},\vy_{g}) = \mathds{1}((\vy_{p} \cap \vy_{g})= \vy_{p}) \\
\text{Comp}(\vy_{p},\vy_{g}) = \mathds{1}((\vy_{p} \cap \vy_{g})= \vy_{g})
\end{align}
\noindent where $\vy_p$ is the predicted and $\vy_g$ is the ground-truth label sets of a given instance, respectively. Overall, the aggregated performance scores for the entire test set are computed as the average of the Jaccard, Accuracy, Correctness, and Completeness scores across all test instances.

\paragraph{Discussion.}
Table~\ref{gap} reveals significant declines in the accuracy of existing MLTC models when transitioning from the iid to the compositional data split for training and evaluation. This performance drop highlights the limited CG capability of MLTC models. Our observations suggest that these models often face issues such as learning spurious correlations or being vulnerable to input perturbations. As seen in emotion classification, the models often predict emotions of the same polarity but struggle to accurately identify compositions that combine emotions of differing sentiment polarities, primarily because such combinations are rare in the training set. We conjecture they have learned a spurious correlation between the frequent co-occurrence of like-polarity emotions. Another challenge for the MLTC models is their inability to accurately predict the same labels corresponding to input texts with notable linguistic variations across the training and test sets. To address these challenges, data augmentation has been shown to mitigate the impact of spurious correlations~\cite{wang2021robustness} and enhance the robustness of machine learning models~\cite{goodfellow2014explaining,huang2021robustness}.


\section{Compositional Data Augmentation}
Compositional data augmentation focuses on sampling examples from the target distribution $P_t(\vx,\vy)$. This distribution can be factorized as $P_t(\vx,\vy)= P_t(\vy)P_t(\vx|\vy)$. The challenge of this approach then becomes to craft two distinct models: one that models the distribution of label compositions $P_t(\vy)$, and another that models the conditional distribution $P_t(\vx|\vy)$. We also introduce a quality control mechanism to filter out low-quality synthetic examples. We hypothesize that exposing MLTC models to diverse syntactic structures and phrases, associated with novel label compositions, can enhance their ability to learn true causal correlations between text and labels. Furthermore, introducing diverse linguistic variations related to each label can improve the model's robustness to input perturbations.






\begin{table}[!t]
\centering
\small
\setlength\tabcolsep{4.3pt} 
\begin{tabular}{lcccccc}
\toprule[0.9pt]
       & \multicolumn{2}{c}{SemEval} & \multicolumn{2}{c}{AAPD} & \multicolumn{2}{c}{IMDB} \\ \cmidrule(r){2-3}\cmidrule(r){4-5}\cmidrule(r){6-7}
Model  & iid     & CG    & iid   & CG   & iid   & CG   \\
\hline
BERT   & 27.31         & 2.85        & 37.79       & 22.02      & 26.17       & 4.35       \\
BERT+P   & 26.88         & 2.71        & 34.22       & 19.08      & 18.93       & 1.95       \\
BERT+MAGNET & 24.95         & 2.23        & 37.70        & 14.48      & 21.73       & 3.68       \\
BERT+SGM    & 19.87             & 2.15           & 37.71      & 15.23          & 18.65           & 3.22          \\
BERT+DBloss & 26.76             & 3.72           & 36.50        & 14.57      & 40.81       & 2.54          \\
T5+CLP & 28.34         & 1.26        & 42.20        & 9.17       & 39.78       & 1.35      \\
\bottomrule[0.9pt]
\end{tabular}
\caption{The classification accuracies of existing MLTC models on both iid and CG splits across three benchmarks.}
\label{gap}
\end{table}

\subsubsection{Label Generative Model.}
To model the label distribution $P_t(\vy)$, we approach it as a sequence modelling task. The learning objective is to optimize the parameters $\vtheta$ to maximize the likelihood of the sequence of tokens in concatenated label phrases within the support set:
\begin{small}
\begin{equation}
\argmax_{\vtheta} \prod_{\vy' \in \mathcal{Y}_{s}} \prod_{t = 0}^{|\vy'|} P_{\vtheta}(y'_t | \vy'_{<t})
\end{equation}
\end{small}

\noindent where $\mathcal{Y}_s$ is the set of all label compositions within the support set $\mathcal{D}_s$, $\vy'$ denotes the sequence of tokens in the concatenated label phrases originating from a label set $\vy$ for each instance, and $y'_t$ is a token at position $t$ in $\vy'$. To accomplish this, we fine-tune a pre-trained language model GPT2~\cite{radford2019language} to estimate the token probability $P_{\vtheta}(y'_t | \vy'_{<t})$, benefiting from its pre-existing knowledge about distributions of label phrase tokens.  In the practical implementation, we prepend a prefix prompt, like `A tweet can express the following emotions:', to the label phrases during training and inference. In the zero-shot setting, where a support set is absent, we instruct GPT2 using the prompt and constrain it only to generate label-related phrases.
\subsection{Conditional Text Generative Models}
As we assume that in the CG split, the source and target conditional distributions align, i.e., $P_s(\vx|\vy) = P_t(\vx|\vy)$, the conditional generation model, therefore, can be trained on the combination of the training and support sets $\mathcal{D}_{cg}$. The learning objective for conditional text generation becomes:
\begin{small}
\begin{equation}
\argmax_{\vtheta} \prod_{\vx,\vy \in \mathcal{D}_{cg}} \prod_{t = 0}^{|\vx|} P_{\vtheta}(x_t | \vx_{<t}, \vy)
\end{equation}
\end{small}

\noindent A common method, as in~\citet{li-etal-2022-variational-autoencoder}, to implement this generation model is to fine-tune a pre-trained language model, such as GPT2 or T5. Pre-trained models appear to excel in compositional sequence generalization~\cite{qiu-etal-2022-evaluating}. During inference, the model converts the concatenated label phrases in the label set $\vy$ into contextualized representations, prompting the generation model to produce text $\vx$ conditioned on the representations. However, after fine-tuned on the compositionally-biased dataset $\mathcal{D}_{cg}$, the representations of different labels, as encoded by the Transformer~\cite{vaswani2017attention}, can be severely entangled. Each label's representation influences others, making it non-invariant to changes in its co-occurring labels. Such entanglement can hinder the generation model from identifying the true associations between each label representation and its corresponding phrases or syntactic structures in the text $\vx$, compromising its ability to composite phrases or structures into high-quality texts for data augmentation. Following this, we present two models focusing on disentangling the label representations for effective text generation.
\subsubsection{Label Specific Prefix-Tuning (LS-PT).}



We suspect that the entanglement observed in label presentations arises from the innate cross-attention mechanism of Transformers. Each label presentation serves as an attended representation of all labels. To address this challenge, we introduce a novel method that encodes each label in the label composition set \(y_i \in \vy\) with its distinct representation, \(\vz_{y_i} \in \mathbb{R}^{L \times H}\). This representation is designed to be minimally influenced by neighboring labels. The concatenated representations of individual labels are processed using a multilayer perceptron (MLP) to produce a composite label representation:
\begin{small}
\begin{equation}
\rmM = \text{MLP} \left( \begin{bmatrix} \vz_{y_0} \\ \vdots \\ \vz_{y_m} \end{bmatrix} \right).
\label{eq:mlp_representation}
\end{equation}
\end{small}

\noindent Building upon the prefix-tuning technique~\cite{li2021prefix}, we estimate the conditional probability \(P_{\vtheta}(x_t | \vx_{<t}, \vy)\), which is further elaborated as $
P_{\vtheta}(x_{t} | \vx_{<t}, \vz_{y_0}, \dots, \vz_{y_m}) = \text{softmax} \left( \vh_{i} \rmW \right).
$ The hidden state \(\vh_i\) is defined as:
\begin{small}
\begin{equation}
\vh_i = 
\begin{cases} 
\rmM[i, :], & \text{if } i \in M_{idx}, \\
\mathbf{LM}(\rmM_{\theta}, \vh_{<i}), & \text{otherwise}.
\end{cases}
\end{equation}
\end{small}

\noindent where the composite label representation \(\rmM \in \mathbb{R}^{ |M_{idx}| \times H'}\) is considered the prefix matrix, \(\vh_i \in \mathbb{R}^{1 \times H'}\) is the hidden state of a language model, \(M_{idx}\) denotes the column indices of the prefix matrix with \(|M_{idx}| = |\vy| \cdot L\), and \(\mathbf{LM}\) refers to a pre-trained auto-regressive language model. Here we adopt GPT2 with frozen parameters as \(\mathbf{LM}\) to maximize benefits from its pre-trained compositional knowledge.

\textit{Inference.} At the inference stage, we draw concatenated label phrases from $P(\vy')$ and convert each set of phrases $\vy'$ into a set of label ids, $\vy$. Texts with new label compositions are then generated conditioned on $\vy$ and label phrases $\vy'$.

\subsubsection{Label Disentangled Variational Autoencoder (LD-VAE).} The model VAE-DPrior~\cite{li-etal-2022-variational-autoencoder} aims to disentangle representations associated with labels and content, and subsequently composite these into new examples. To realize this, the model adjusts its estimation from $P_{\vtheta}(\vx | \vy)$ to $P_{\vtheta}(\vx | \vy,c)$, where $c$ denotes a variable capturing the prior knowledge of content related to the text $\vx$. The learning objective of VAE-DPrior is to maximize the Evidence Lower Bound (ELBO) of $P_{\vtheta}(\vx | \vy,c)$. Specifically, after introducing a latent variable $\vz_c$ associated with the content $c$ and a latent variable $\vz_\vy$ associated the label $\vy$, and providing a strong conditional independence assumption, $P(\vz_c, \vz_\vy | \vz, c, \vy) = P(\vz_c | \vx, c) P(\vz_\vy | \vx, \vy)$, ELBO objective for VAE-DPrior is:
\begin{align}
\label{eq:elbo}
    \begin{split}
        &\mathbb{E}_{Q_{\vPhi}(\vz_c, \vz_\vy | \vx, c, \vy)}[\log P_{\vTheta}(\vx| \vz_c, \vz_\vy )]\ \ \}\gL_r \\
        &- \KL(Q_{\vPhi}(\vz_c | \vx, c)  \| P_{\vTheta}(\vz_c | c))\ \ \ \ \ \ \}\gL_c  \\ 
        &- \KL(Q_{\vPhi}(\vz_\vy | \vx, \vy)  \| P_{\vTheta}(\vz_\vy | \vy))\ \ \ \ \}\gL_l
    \end{split}
\end{align}

\noindent where $\gL_r$ represents the text reconstruction loss for the VAE decoder, while $\gL_c$ and $\gL_l$ are the regularization loss terms for content and label encoders, respectively. With sufficiently divergent prior conditional distributions $P_{\vTheta}(\vz_c | c)$ and $P_{\vTheta}(\vz_\vy | \vy)$ during regularization, VAE-DPrior disentangles label and content representations. However, the model overlooks disentanglement within label representations. To address this, we consider a set of representations $\vz_{\vy} = \{\vz_{y_0},...,\vz_{y_m}\}$, where each label $y$ from the label set $\vy$ is \textit{only} associated with a specific latent variable $\vz_{y}$. Adopting a conditional independence assumption, given by $P(\vz_\vy | \vx, \vy) =  \prod_{i = 0}^{m} P(\vz_{y_{i}} | \vx, y_{i})$ and $P(\vz_\vy | \vy) =  \prod_{i = 0}^{m} P(\vz_{y_{i}} | y_{i})$, we update $\gL_l$ using the chain rule of the Kullback–Leibler (KL) divergence:
\begin{small}
\begin{equation}
    \gL_l = - \sum_{i = 0}^{m} \KL(Q_{\vPhi}(\vz_{y_{i}} | \vx, y_{i})  \| P_{\vTheta}(\vz_{y_{i}} | y_i))
\end{equation}
\end{small}

\noindent We aim to disentangle the label representations by employing distinct conditional priors for different label encoders. Since our focus is not on the deep theoretical foundations of disentanglement learning, those interested in the theory can refer to the original VAE-DPrior work. The proof of how we leverage the KL chain rule for $\gL_l$ can be found in Appendix. Next, we delve into the implementation of LD-VAE.
\begin{table*}[!t]
\centering
\small
\label{main_results}
\begin{tabular}{lcccccccccccc}
\toprule[0.9pt]
& \multicolumn{4}{c}{SemEval}                                         & \multicolumn{4}{c}{AAPD}                                            & \multicolumn{4}{c}{IMDB}                                                                 \\ \cmidrule(r){2-5}\cmidrule(r){6-9}\cmidrule(r){10-13}
Model                    & \multicolumn{1}{c}{Jacc}    & \multicolumn{1}{c}{Acc}    & \multicolumn{1}{c}{Corr}   & Comp   & \multicolumn{1}{c}{Jacc}   & \multicolumn{1}{c}{Acc}    & \multicolumn{1}{c}{Corr}   & Comp   & \multicolumn{1}{c}{Jacc}   & \multicolumn{1}{c}{Acc}    & \multicolumn{1}{c}{Corr}   & Comp   \\ \hline
\textit{No Aug.}   & \multicolumn{1}{c}{44.90} & \multicolumn{1}{c}{2.92} & \multicolumn{1}{c}{49.23} & 15.31 & \multicolumn{1}{c}{52.24}  & \multicolumn{1}{c}{21.96}  & \multicolumn{1}{c}{42.38}  & 42.54  & \multicolumn{1}{c}{42.94}  & \multicolumn{1}{c}{4.48}   & \multicolumn{1}{c}{61.09}  & 8.77   \\ 
\hline
Concat                   & \multicolumn{1}{c}{45.84}   & \multicolumn{1}{c}{3.25}  & \multicolumn{1}{c}{48.37} & 15.14 & \multicolumn{1}{c}{-}      & \multicolumn{1}{c}{-}      & \multicolumn{1}{c}{-}      & -      & \multicolumn{1}{c}{46.13} & \multicolumn{1}{c}{8.71}  & \multicolumn{1}{c}{61.02} & 14.25 \\ 
Flan-T5                  & \multicolumn{1}{c}{47.84}  & \multicolumn{1}{c}{8.01}  & \multicolumn{1}{c}{52.53} & \textbf{17.32} & \multicolumn{1}{c}{55.98} & \multicolumn{1}{c}{26.46}  & \multicolumn{1}{c}{44.42} & 45.68 & \multicolumn{1}{c}{47.39}  & \multicolumn{1}{c}{11.69} & \multicolumn{1}{c}{60.70} & 16.98 \\ 
VAE-DPrior                   & \multicolumn{1}{c}{47.50}       & \multicolumn{1}{c}{6.07}      & \multicolumn{1}{c}{49.62}      & 17.17      & \multicolumn{1}{c}{55.52}      & \multicolumn{1}{c}{24.76}      & \multicolumn{1}{c}{42.28}      & 46.30      & \multicolumn{1}{c}{46.49}      & \multicolumn{1}{c}{9.68}      & \multicolumn{1}{c}{59.29}      & 15.54      \\ 
GPT3.5                  & \multicolumn{1}{c}{46.56}       & \multicolumn{1}{c}{4.74}      & \multicolumn{1}{c}{47.58}      & 16.69      & \multicolumn{1}{c}{56.52}      & \multicolumn{1}{c}{26.63}      & \multicolumn{1}{c}{44.77}      & 46.42      & \multicolumn{1}{c}{46.69}      & \multicolumn{1}{c}{10.04}      & \multicolumn{1}{c}{62.85}      & 15.13      \\ 
GPT2-PT            & \multicolumn{1}{c}{47.44}  & \multicolumn{1}{c}{6.81}  & \multicolumn{1}{c}{50.53} & 17.22 & \multicolumn{1}{c}{56.38}  & \multicolumn{1}{c}{28.03}  & \multicolumn{1}{c}{45.95}  & 46.22  & \multicolumn{1}{c}{47.12}  & \multicolumn{1}{c}{11.38}  & \multicolumn{1}{c}{61.48}  & 16.35 \\ 
\hline
LS-PT & \multicolumn{1}{c}{\textbf{48.02}}   & \multicolumn{1}{c}{8.29}  & \multicolumn{1}{c}{52.94} & 16.93 & \multicolumn{1}{c}{57.95} & \multicolumn{1}{c}{30.21} & \multicolumn{1}{c}{47.74} & 47.23 & \multicolumn{1}{c}{\textbf{48.36}}     & \multicolumn{1}{c}{\textbf{12.56}} & \multicolumn{1}{c}{62.37} & \textbf{17.57} \\ 
LD-VAE                      & \multicolumn{1}{c}{47.94}  & \multicolumn{1}{c}{\textbf{8.44}}  & \multicolumn{1}{c}{\textbf{53.10}}  & 16.96  & \multicolumn{1}{c}{\textbf{58.50}} & \multicolumn{1}{c}{\textbf{31.11}} & \multicolumn{1}{c}{\textbf{48.67}}  & \textbf{48.09} & \multicolumn{1}{c}{48.07} & \multicolumn{1}{c}{11.75}  & \multicolumn{1}{c}{\textbf{63.23}}  & 16.92 \\ 
\bottomrule[0.9pt]

\end{tabular}
\caption{Classification results using BERT with augmentation instances from various generators on the CG data split.}
\label{augmentation_results}
\end{table*}

\textit{Regularization for Content Encoders.} In line with VAE-DPrior, the content knowledge $c \in \mathcal{C}$ derived from an input text $\vx$ is represented by one of the $|\mathcal{C}|$ centroids. These centroids are formed using $k$-means clustering, with BERT~\cite{bert} encoding each text $\vx \in \mathcal{X}_{cg}$ in $\mathcal{D}_{cg}$. Formally, the content prior $P_{\vTheta}(\vz_c | c)$ assumes the form of a conditional Gaussian, $\mathcal{N}(\vz_c ; \vMu^p_{c}, \lambda_c\rmI)$. Here, the mean $\vMu^p_{c} =  \vk_{c}\rmW_c$ is a linear projection  of the relevant cluster centroid vector $\vk_c \in \mathbb{R}^{1 \times H}$. The text $\vx$ belongs to the cluster with centroid $c$. $I$ is an identity matrix determining variance.

The posterior distribution $Q_{\vPhi}(\vz_c | \vx, c)$ is modeled by the \textbf{content encoder}, using the VAE reparameterization trick:
\begin{small}
\begin{align}
\label{eq:content}
 \vv_c &= \text{Mean}(\text{GRU}_{c}(\rmV_\vx)) \\
    \log \bm{\sigma}^q_c, \vMu^q_c &= \text{MLP}_{\bm{\sigma}^q_c}(\vv_c), \text{MLP}_{\vMu^q_c}(\vv_c)  \\
    \vz_c &= \vMu^q_c + \exp(\frac{1}{2}\log\bm{\sigma}^q_c) \odot \bm{\epsilon}_c
\end{align}
\end{small}

\noindent where $\odot$ is the element-wise product, $\bm{\epsilon}_c$ is Gaussian noise from the distribution $\mathcal{N}(0,\lambda_c\rmI)$. $\rmV_\vx$ is the contextualized representation of $\vx$ encoded by BERT with parameters frozen, the GRU is Gated Recurrent Units~\cite{cho2014learning}, and $\vz_c \in \mathbb{R}^{1 \times H}$ is a one-dimensional vector. Ideally, each variable 
$c$ would require separate parameters, leading to $|\mathcal{C}|$ content encoders and prior conditional Gaussians with distinct sets of parameters. Given the large size of $\mathcal{C}$, we utilize shared parameters from GRUs, and MLPs across all $c \in \mathcal{C}$ for parameter efficiency. Hence, the content encoder's regularization loss $- \KL(Q_{\vPhi}(\vz_c | \vx, c)  \| P_{\vTheta}(\vz_c | c))$ is formulated as maximizing:
\begin{small}
\begin{equation}
\mathcal{L}_c = -\frac{1}{2\lambda_c}\left( \|\vMu_c^q - \vMu_c^p\|^2 +  (\vSigma_c^q - \lambda_c \log(\vSigma_c^q))\cdot \mathbf{1} \right)
\end{equation}
\end{small}
\noindent where $\mathbf{1} \in \mathbb{R}^{H \times 1}$ is used to sum over elements of the vector.

\textit{Regularization for Label Encoders.} For each label $y_i \in \vy$, the conditional prior $P_{\vTheta}(\vz_{y_i} | y_i)$ takes the form of a conditional Gaussian form, represented as $\mathcal{N}(\vz_{y_i} ; \vMu^p_{y_i}, \lambda_{y_i}\rmI)$. The mean of this Gaussian corresponds to a linear projection of the embedding of the label phrase, denoted as $\vMu^p_{y_i} = \vl_{y_i} \rmW_{y_i}$, with $\vl_{y_i}$ being encoded using frozen BERT.

The posterior distribution $Q_{\vPhi}(\vz_{y_i} | \vx, y_i)$ is modeled by \textbf{label encoder}, which mirrors the structure of the content encoder. Unlike the content encoder, the GRU parameters are not shared. Therefore, we apply $m$ label encoders with distinct sets of parameters, with each set modeling the posterior distribution for one latent variable $\vz_{y_i}$.


The regularization loss for the label encoders becomes:
\begin{small}
\begin{equation}
\mathcal{L}_l = \sum_{i=1}^{m} -\frac{1}{2\lambda_{y_i}}\left( \|\vMu_{y_i}^q - \vMu_{y_i}^p\|^2 + (\vSigma_{y_i}^q - \lambda_{y_i} \log(\vSigma_{y_i}^q))\cdot \mathbf{1} \right)
\end{equation}
\end{small}

\textit{Text Reconstruction.}  The reconstruction loss is the maximum likelihood loss used to optimize the parameters of the decoder. The decoder shares the same structure as LS-PT and employs prefix-tuning, allowing a GPT2 to generate text conditioned on $\vz_c$ and $\{\vz_{y_0},...,\vz_{y_m}\}$. There is a slight difference from Eq.~\ref{eq:mlp_representation} as shown below:
\begin{small}
\begin{equation}
\rmM = \text{MLP} \left( \begin{bmatrix} \vz'_{y_0}; \vz'_c \\ \vdots \\ \vz'_{y_m}; \vz'_c \end{bmatrix} \right)
\label{eq:mlp_representation_vae}
\end{equation}
\end{small}

\noindent where $\vz'_{y_i} \in \mathbb{R}^{L \times H}, \vz'_{c} \in \mathbb{R}^{L \times H}$ is obtained by repeating $\vz_{y_i}$ and $\vz_{c}$ $L$ times, respectively, because we expect a longer prefix length can enhance the performance of prefix-tuning. The $\vz'_{y_i}$ and $\vz'_c$ are vertically concatenated together.


\textit{Inference.} During inference, the encoders are discarded. The content variable $c$ is randomly sampled from $\mathcal{C}$, based on the assumption that $P(c)$ follows a uniform distribution. The label set $\vy = \{y_0,...,y_m \}$ is drawn from the label generative model $P(\vy')$. With the sampled $c$ and $\vy$, we sample latent representations $\vz_c$ and $\vz_{y_i}$ from the conditional Gaussian priors $P_{\vTheta}(\vz_c | c)$ and $P_{\vTheta}(\vz_{y_i} | y_i)$, respectively. The decoder then generate synthetic text $\vx'$ conditioned on the latent label representations $\vy$ and label phrases $\vy'$.
\subsubsection{Quality Control (QC).} QC is implemented using a BERT-based MLTC classifier trained on $\mathcal{D}_{cg}$. We first overgenerate synthetic examples, each text $\vx$ paired with a label set $\vy_s$. Next, we use the classifier to predict the labels $\vy_p$ for each text $\vx$. We then rank the examples by their Jaccard scores, $\text{Jacc}(\vy_p,\vy_s)$, and retain those with the top $K$ highest scores.

\section{Experiments}

\subsubsection{Datasets.}
We conduct experiments on the compositional splits of three datasets: SemEval~\cite{semeval}, AAPD~\cite{SGM}, and IMDB~\cite{IMDB}. During the data splitting process, we allocate 20 label compositions to the test set. After splitting, SemEval, a multi-label emotion classification dataset, comprises 9,530 training, 50 support, and 1,403 test examples. AAPD features academic paper abstracts annotated with subject categories from Arxiv and contains 50,481 training, 50 support, and 5,309 testing examples. IMDB provides movie reviews annotated with movie genres and includes a total of 107,944 training, 50 support, and 9,200 test samples. We first overgenerate 2,000, 10,000, and 24,000 examples, and then apply quality control to filter the synthetic data down to sizes of 1,000, 5,000, and 12,000 for SemEval, AAPD, and IMDB, respectively. 

\subsubsection{MLTC Models.}
We evaluate the performance of MLTC models on our CG tasks:
\textit{i)} \textbf{BERT~\cite{bert}} employs BERT combined with MLPs on BERT's top layers for multi-label classification and is optimized using cross-entropy loss.
\textit{ii)} \textbf{BERT+P} has the same structure as \textbf{BERT} but is optimized using p-tuning~\cite{p-tuning}.
\textit{iii)} \textbf{BERT+DBloss~\cite{dbloss}} uses BERT and is optimized with a loss function tailored specifically to address label distribution imbalances in MLTC datasets.
\textit{iv)} \textbf{BERT+MAGNET~\cite{magnet}} integrates BERT with a graph attention network designed to learn correlations between labels.
\textit{v)} \textbf{BERT+SGM~\cite{SGM}} treats MLTC as a sequence generation task. The word embeddings of the text, encoded by BERT, are then fed into an LSTM that learns label correlations and generates label predictions.
\textit{vi)} \textbf{T5+CLP~\cite{clp}} is a model based on T5~\cite{raffel2020exploring} designed to capture label correlations using the decoder combined with a contrastive learning loss.
\begin{table*}[!ht]
\centering
\small
\begin{tabular}{lcccccccccccc}
\bottomrule[0.9pt]

Model     & \multicolumn{2}{c}{BERT}  & \multicolumn{2}{c}{BERT+P}       & \multicolumn{2}{c}{BERT+DBloss}               & \multicolumn{2}{c}{BERT+MAGNET}               & \multicolumn{2}{c}{BERT+SGM} & \multicolumn{2}{c}{T5+CLP}                    \\ \cmidrule(r){2-3}\cmidrule(r){4-5}\cmidrule(r){6-7}\cmidrule(r){8-9}\cmidrule(r){10-11}\cmidrule(r){12-13}
{}     & \multicolumn{1}{c}{-}  & \multicolumn{1}{c}{+}  & \multicolumn{1}{c}{-}  & \multicolumn{1}{c}{+} & \multicolumn{1}{c}{-} & \multicolumn{1}{c}{+} & \multicolumn{1}{c}{-} & \multicolumn{1}{c}{+} & -             & +            & \multicolumn{1}{c}{-} & \multicolumn{1}{c}{+} \\
\hline
{Jacc}     & 52.24 & 58.50      & 49.16 & 53.65           & 50.92                 & 58.52                 & 42.26                 & 51.27                 & 43.95             & 53.37            & 43.03                 & 43.72                 \\
Acc         & 21.96    & 31.11       & 19.06 & 28.98          & 17.00                 & 27.02                 & 15.07                 & 30.02                 & 15.77             & 28.64            & 9.32                  & 15.89                 \\
Corr      & 42.38    & 48.67     & 40.86 & 49.58            & 30.18                 & 35.35                 & 43.22                 & 52.57                 & 41.65             & 51.98            & 14.23                 & 20.96                 \\
Comp     & 42.54    & 48.09      & 34.02 & 42.05           & 45.28                 & 59.00                 & 25.01                 & 36.72                 & 28.96             & 37.23            & 34.99                 & 33.52                             
\\
\bottomrule[0.9pt]
\end{tabular}
\caption{The results of different multi-label text classifiers on the AAPD dataset. The symbols ``+'' and ``-'' denote whether the augmentation data generated by LD-VAE is utilized during classifier training or not, respectively.}

\label{classifiers}
\end{table*}

\subsubsection{Generator Baselines.}
We compare five conditional text generators trained on $\mathcal{D}_{cg}$. Each generator generates text conditioned on the same set of novel label compositions, which are sampled from the label generative model. Furthermore, all employ the same filter model for quality control.
\textit{i)} \textbf{Concat.} This method simply concatenates single-labeled instances to create synthetic examples with specific label compositions. This concept aligns with the approach taken by~\citet{jia2016data} for the semantic parsing task. Note: AAPD is not suited for this baseline because it lacks single-labeled instances.
\textit{ii)} \textbf{Flan-T5~\cite{FLAN-T5}} is a sequence-to-sequence language model~\cite{sutskever2014sequence} pre-trained on thousands of NLP tasks. It crafts text based on composites of label phrases processed by its encoder.
\textit{iii)} \textbf{VAE-DPrior~\cite{li-etal-2022-variational-autoencoder}} employs the VAE and disentanglement learning to disentangle label and content representations, and then generates new texts conditioned on a combination of these representations.
\textit{iv)} \textbf{GPT3.5}\footnote{https://chat.openai.com/}: We direct GPT3.5 to produce texts based on concatenated label phrases. Moreover, we deploy few-shot in-context learning, enabling GPT3.5 to observe all existing examples labeled with the respective compositions from the label generative model.
\textit{v)} \textbf{GPT2-PT~\cite{li2021prefix}}: This approach fine-tunes GPT2 using prefix-tuning, generating text that begins with the concatenated label phrases.


\subsection{Main Results and Discussions.}
\paragraph{Analysis of Generators.} Table~\ref{augmentation_results} shows that utilizing synthetic data, generated by \textit{all} types of generators, enhances the performance of the BERT classifier on the CG splits of three benchmarks across various evaluation metrics. This evidence underscores the effectiveness of our data augmentation approach in addressing the CG challenge within MLTC. Both LS-PT and LD-VAE outperform the other baselines, highlighting the essential role of disentangled label representations for generating high-quality instances with novel label compositions. In contrast, the Concat baseline underperforms, likely because the concatenated text is neither semantically nor syntactically coherent. Flan-T5 and GPT2-PT produce text based on label representations encoded via Transformer layers. However, we believe their encoding methods may result in entangled label representations, which may explain their inferior performance in data augmentation compared to our method. While VAE-DPrior adopts disentanglement learning and latent label representations, its lack of a label-specific representation for each label makes it less directly comparable to our approach. Even though GPT3.5 is recognized as a powerful language model, it does not excel in augmenting the CG abilities of MLTC models, potentially because it is exposed to only a few-shot examples. It's worth noting that AAPD excludes single-labeled instances, making learning disentangled label representations challenging. Yet, LD-VAE still excels on AAPD, whereas Flan-T5, despite its strong performance on the other two datasets, falls short. 

\paragraph{Analysis of Classifiers.} Table~\ref{classifiers} shows that all MLTC models evaluated struggle with CG, each for unique reasons.  \citet{qiu-etal-2022-evaluating} found that models using parameter-efficient fine-tuning (PEFT) generally outperform merely fine-tuned ones in out-of-distribution CG scenarios in semantic parsing, a natural language understanding (NLU) task. However, BERT+P, despite employing PEFT, does not outperform fine-tuned baseline in MLTC — a task also under NLU. MLTC models designed to learn label correlations, like BERT+MAGNET, BERT+SGM, and T5+CLP, score better in correctness but have lower completeness scores than other baselines, suggesting they tend to predict only a subset of the ground truth. Interestingly, despite T5+CLP achieving several state-of-the-art results on current MLTC benchmarks with the standard data split, it performs the worst among all baselines. We conjecture that this line of work, despite its popularity, might produce models prone to learning spurious correlations among labels. In contrast, BERT+DBloss, designed to tackle label imbalance, leans towards over-predicting labels with its high completeness score. We also investigate the impact of synthetic data, generated by LD-VAE, on the performance of these models. 
Incorporating this synthetic data into training significantly boosts \textit{all} models regarding the evaluation metrics, demonstrating the effectiveness of our data augmentation strategy in helping various MLTC models address CG challenges.

\subsection{Ablation Study}

\subsubsection{Support Size.} We investigate the influence of the sizes of support set on three aspects when: \textit{i)} fine-tuning the label generator, \textit{ii)} learning the conditional text generator, and \textit{iii)} training the classifier.
In each experiment in Table~\ref{sample_size}, we fix the support set data size for the other two aspects at 50 and only vary the data size for one aspect at a time.
All experiments share the same quality control filter and test set for fair comparisons.
Key takeaways include: \textit{i)} With just 50 sampled examples, the label generator can estimate a label composition distribution reasonably close to what is achieved with 250 examples. However, a zero-shot approach that relies solely on the pre-trained knowledge of label token distribution remains challenging, resulting in the classifier accuracy being about 10\% lower than using 50 examples. \textit{ii)} Enhancing the conditional generator with additional support data has minimal impact on MLTC performance, given that even 250 examples occupy just a small fraction of the overall training data. This further solidifies our hypothesis that the conditional distribution does not shift across CG splits. 
\textit{iii)} Support data size crucially affects classifier training. More human-crafted data improves classifier performance in CG.

\begin{table}[!t]
\centering
\small
\begin{tabular}{lccc}
\toprule[0.9pt]
                       \makecell[l]{Size of \\ Support}       & \makecell[c]{Label \\Generator} & \makecell[c]{Conditional \\ Text Generator} & \makecell[c]{Classifier} \\
                       \hline
0           & 23.56                  & 31.41                   & 32.37                \\
50          & 32.09              & 32.09               & 32.09            \\
100         & 30.02                  & 31.70                   & 33.31            \\
250         & 31.61                  & 31.92                   & 35.23   \\
\bottomrule[0.9pt]

\end{tabular}
\caption{Accuracies of the BERT classifier on AAPD, across three modules, with varying support data sizes, using LD-VAE as the conditional text generator.}
\label{sample_size}
\end{table}




\begin{table}[!t]
\small
\centering
\begin{tabular}{lcccc}
\toprule[0.9pt]
                & \multicolumn{2}{c}{SemEval}                & \multicolumn{2}{c}{AAPD}                   \\
                \cmidrule(r){2-3}\cmidrule(r){4-5}
                & \multicolumn{1}{c}{filter} & random & \multicolumn{1}{c}{filter} & random \\
                \hline
Flan-T5         & 8.01                         & 6.18      & 26.46                         & 24.01         \\
GPT2-PT   & 6.81                         & 6.63       & 28.03                         & 27.16      \\
\hline
\makecell[c]{LS-PT} & 8.29                         & 6.93      & 30.21                        & 29.32      \\
LD-VAE             & 8.44                         & 6.43      & 31.11                        & 28.82   \\
\bottomrule[0.9pt]
\end{tabular}
\caption{
Accuracies of BERT classifiers with and without filtering synthetic data generated by various generators.}
\label{filter}
\end{table}

\subsubsection{Quality Control.}
We investigate the effectiveness of QC by comparing it with random selection. The sizes of selected synthetic data are equal for both settings.
As shown in Table~\ref{filter}, our BERT-based filter improves the quality of the generated examples, as evidenced by the higher accuracy of the classifier trained on filtered data.
We note that the filter tends to discard low-quality synthetic examples that are either irrelevant to the target label composition or texts with no practical meaning, such as Twitter tags and blank text.

\subsubsection{Disentanglement.} 
\begin{figure}[!t]
    \centering
    \includegraphics[width=0.65\linewidth]{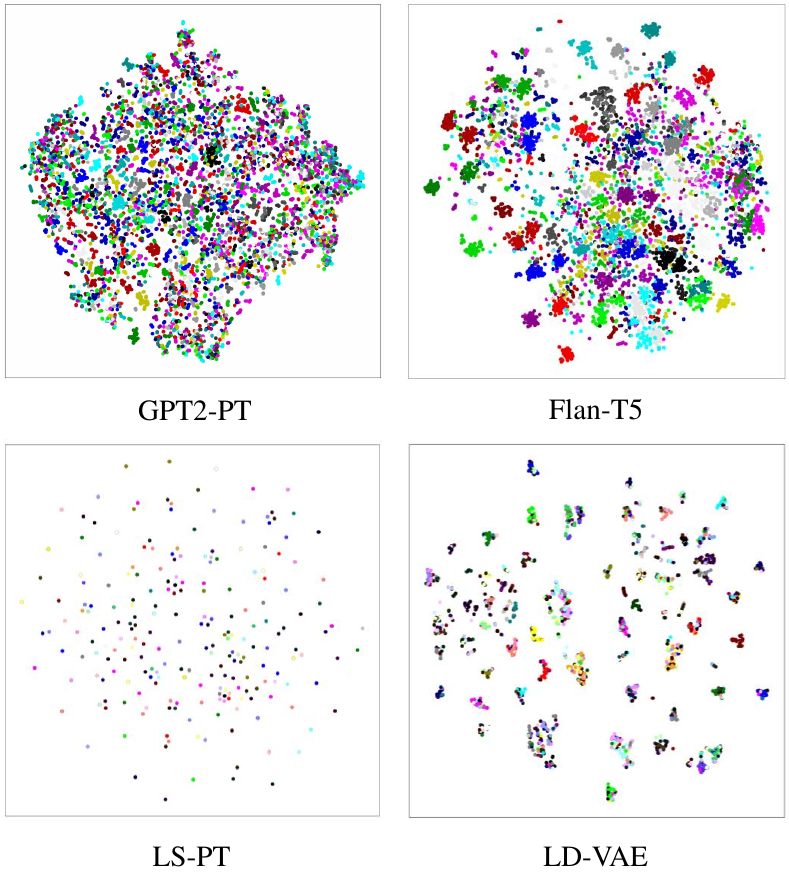}
    \caption{T-SNE visualization of Transformer-encoded label phrase representations from GPT2-PT and Flan-T5 versus latent label representations in the prefixes of LS-PT and LD-VAE. Each label, within varying label compositions from the training set $\mathcal{D}_{cg}$ of AAPD, is represented by a distinct colour.
    }
    \label{disentangle}
\end{figure}

Figure~\ref{disentangle} shows entangled representations of label phrases from GPT2-PT. In contrast, the label phrase representations encoded by Flan-T5 remain invariant regardless of label composition changes, with representations of the same labels clustering closely. This may be due to Flan-T5's unique pre-training method across thousands of NLP tasks, allowing it to encounter diverse text compositions. This unique training method may explain why Flan-T5 outperforms most of the baselines. On the other hand, our methods disentangle latent representations more effectively than both GPT2-PT and Flan-T5. Notably, LD-VAE samples representations from a more continuous space rather than focusing on a singular point for each label, resulting in a more cohesive and fluent generated text than LS-PT, given our manual inspection. A further experiment reveals that replacing the label-conditioned priors in our LD-VAE with a normal distribution, as seen in vanilla VAEs, leads to a 5\% drop in BERT classifier accuracy on AAPD. This emphasizes the significance of disentanglement learning.



\section{Related Work}
\paragraph{Multi-label Text Classification.}


In the field of MLTC, studies address the critical challenge of label correlation through methods such as incorporating label co-occurrence~\cite{magnet,10.1007/978-3-030-60450-9_51} and employing correlation loss functions~\cite{clp,spanemo}. Some studies also adopt sequence-to-sequence approaches for MLTC, wherein the decoder takes label correlations into account~\cite{SGM,huang-etal-2021-seq2emo}. Beyond label correlation, several works employ attention mechanisms to incorporate contextual label information during the prediction~\cite{xiao-etal-2019-label,huang2021label}. Additionally, various works address the challenge of label distribution imbalance in MLTC~\cite{dbloss,yang-etal-2020-hscnn,cao2019learning}. However, these studies mainly deal with the scarcity of individual labels. In contrast, our focus is on the datasets where individual labels are well-represented, but certain label combinations remain sparse.


\paragraph{Compositional Generalization.}
CG has been explored in various NLP domains, including semantic parsing~\cite{qiu2022improving,andreas2020good,yang2022subs,qiu-etal-2022-evaluating,haroutunianreranking}, controllable text generation~\cite{li-etal-2022-variational-autoencoder,zeng2023seen},
single-label classification~\cite{li-etal-2022-variational-autoencoder}, and machine translation~\cite{li2021compositional,russin2019compositional,zheng2022disentangled}. Typically, these studies enhance the CG capabilities of models in their respective tasks using methods such as data augmentation~\cite{jia2016data,andreas2020good,qiu2022improving}, leveraging pre-trained knowledge from language models~\cite{qiu-etal-2022-evaluating,furrer2020compositional}, employing disentanglement learning~\cite{zheng2022disentangled,montero2020role} for improved latent representations, or a hybrid approach~\cite{li-etal-2022-variational-autoencoder}, similar to ours.


\section{Conclusion}
In summary, we examined the CG challenges in current MLTC models using our unique evaluation metrics and data splits. Our findings reveal a significant deficit in their CG capabilities, limiting their generalization to rare compositional concepts. To address this, we introduced a data augmentation method paired with two conditional text generators that learn disentangled label representations, enabling higher-quality text generation. Empirical results demonstrate that our method significantly mitigates the CG issue for MLTC models, with our generators surpassing other baseline counterparts in enhancing CG capabilities of these models.

\section{Acknowledgments}
This work is supported by the National Natural Science Foundation of China (No. 62176187).

\bibliography{aaai24}

\newpage
\appendix
\onecolumn
\section{Appendix}
\subsection{A\hspace{10pt}Implementation Details}
We set the learning rate at 5e-5 for both LS-PT and LD-VAE. For classifier training, BERT uses a rate of 2e-5, while other modules are at 1e-3. We reserve 5\% of the training data for validation to obtain the hyperparameters. All tests average results over five runs, each with a unique seed. Both BERT-base (with 110M parameters) and GPT2 (also with 110M parameters) serve as foundational models — BERT for LS-PT, LD-VAE, VAE-DPrior and all BERT-based MLTC models, and GPT2 for label generation and decoders in LS-PT, LD-VAE, VAE-DPrior and GPT2-PT. Every model is trained and evaluated on a single RTX 4090 GPU. The classifier T5+CLP utilizes a T5 of 220 million in size. Meanwhile, the Flan-T5 text generator has 250 million parameters, aligning closely with the sizes of the other two sequence-to-sequence models, LD-VAE and VAE-DPrior.

\subsection{B\hspace{10pt}Evaluation Metrics}
Let $\vy_p$ denote the predicted label set and $\vy_g$ denote the ground-truth label set. The \textbf{Jacc}ard index score and exact match \textbf{Acc}uracy are defined as:
\begin{align}
\text{Jacc}(\vy_{p},\vy_{g}) = \frac{|\vy_{p} \cap \vy_{g}|}{|\vy_{p} \cup \vy_{g}|}.
\end{align}

\begin{align}
\text{Acc}(\vy_{p},\vy_{g}) = \mathds{1}(\vy_{p}=\vy_{g}).
\end{align}

\subsection{C\hspace{10pt}Data Statistics}
\begin{table}[!ht]
\centering
\begin{tabular}{lccccccccc}
\toprule[0.9pt]
        & $N_{train}$ & $N_{test}$& $N_{sup}$ & $\overline{W}$     & $Y$  & $\overline{Y}$   & $Y_{min}$ & $Y_{max}$ & $M$  \\
        \hline
SemEval & 9,530   & 1,403 & 50  & 16.0    & 11 & 2.4 & 0    & 6    & 20 \\
AAPD    & 50,481  & 5,309 & 50  & 163.6 & 54 & 2.4 & 2    & 8    & 20 \\
IMDB    & 107,944 & 9,200 & 50  & 98.4  & 27 & 2.2 & 1    & 12   & 20 \\
\bottomrule[0.9pt]
\end{tabular}
\caption{Details of our CG split dataset. $N_{train}$, $N_{test}$, and $N_{sup}$ represent the number of training, test, and support instances, respectively. $\overline{W}$ indicates the average word count per input, while $Y$ is the total class count. $\overline{Y}$ represents the average number of labels for each label composition. $Y_{min}$ and $Y_{max}$ denote the minimum and maximum label counts within individual label compositions in the dataset. Lastly, $M$ specifies the number of label compositions in the test set.}
\label{dataset}
\end{table}

We conduct experiments using the SemEval, AAPD, and IMDB MLTC datasets. Table~\ref{dataset} presents the details of our dataset in the CG split. SemEval is a multi-label emotion classification dataset with 10,983 tweets. We filter out instances without any labels during generator training. AAPD contains 55,840 academic paper abstracts annotated with their subjects from Arxiv. IMDB comprises 117,194 movie reviews annotated with movie genres. We randomly select specific label compositions, and instances labelled with these compositions are considered as the test set.

\subsection{D\hspace{10pt}KL Chain Rule for $\gL_l$}
\begin{figure}[!ht]
    \centering
    \includegraphics[width=0.2\linewidth]{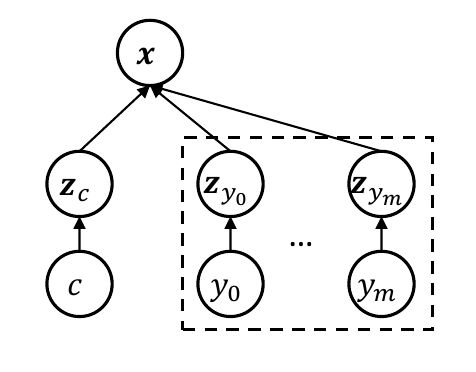}
    \caption{ The Bayesian graph for the LD-VAE.
    }
    \label{proof}
\end{figure}

\label{app:chain_rule}
Based on the original proof presented in Section E3.2 of the work that presents VAE-DPrior~\cite{li-etal-2022-variational-autoencoder}, the assumption of conditional independence is formulated as $P(\vz_c, \vz_\vy | \vx, c, \vy) = P(\vz_c | \vx, c) P(\vz_\vy | \vx, \vy)$. Given this assumption, the evidence lower bound (ELBO) of $P_{\vTheta}(\vx|\vy,c)$ is:
\begin{align}
    P_{\vTheta}(\vx|\vy,c) \geq & \mathbb{E}_{Q_{\vPhi}(\vz_c, \vz_\vy | \vx, c, \vy)}[\log P_{\vTheta}(\vx| \vz_c, \vz_\vy )] - \KL(Q_{\vPhi}(\vz_c | \vx, c)  \| P_{\vTheta}(\vz_c | c)) - \KL(Q_{\vPhi}(\vz_\vy | \vx, \vy)  \| P(\vz_\vy | \vy)).
\end{align}

Given the label set $\vy = \{y_0,...,y_m\}$ and the corresponding latent variables $\vz_\vy = \{\vz_{y_0},...,\vz_{y_m}\}$, the following relationship $\KL(Q_{\vPhi}(\vz_\vy | \vx, \vy)  \| P_{\vTheta}(\vz_\vy | \vy)) = \sum_{i = 0}^{m} \KL(Q_{\vPhi}(\vz_{y_{i}} | \vx, y_{i})  \| P_{\vTheta}(\vz_{y_{i}} | y_i))$ holds under the assumption of conditional independence, which suggests: $P(\vz_\vy | \vx, \vy) =  \prod_{i = 0}^{m} P(\vz_{y_{i}} | \vx, y_{i})$ and $P(\vz_\vy | \vy) =  \prod_{i = 0}^{m} P(\vz_{y_{i}} | y_{i})$, as described by the Bayesian graph for the LD-VAE in Figure~\ref{proof}.

\noindent
\textit{Proof:}

\begin{align}
     &\KL(Q_{\vPhi}(\vz_\vy | \vx, \vy) \| P_{\vTheta}(\vz_\vy | \vy)) \\
    =& \int Q_{\vPhi}(\vz_{\vy} | \vx, \vy) \Big[\log \frac{P_{\vTheta}(\vz_{\vy}|\vy))}{Q_{\vPhi}(\vz_\vy | \vx, \vy)} \Big] d\vz_\vy \\
    =& \int Q_{\vPhi}(\vz_{y_0},...,\vz_{y_m} | \vx, y_0,...,y_m) \Big[\log \frac{P_{\vTheta}(\vz_{y_0},...,\vz_{y_m}|y_0,...,y_m)}{Q_{\vPhi}(\vz_{y_0},...,\vz_{y_m} | \vx, y_0,...,y_m)} \Big] d\vz_{y_0},...,d\vz_{y_m} \\
    =& \int \prod_{i = 0}^{m} Q_{\vPhi}(\vz_{y_{i}} | \vx, y_{i}) \Big[\log \frac{\prod_{i = 0}^{m} P_{\vTheta}(\vz_{y_{i}} | y_{i})}{\prod_{i = 0}^{m} Q_{\vPhi}(\vz_{y_{i}} | \vx, y_{i})} \Big] d\vz_{y_0},...,d\vz_{y_m} \\
    =& \sum_{i = 0}^{m} \int Q_{\vPhi}(\vz_{y_i} | \vx, y_i) \Big[\log \frac{P_{\vTheta}(\vz_{y_i}|y_i)}{Q_{\vPhi}(\vz_{y_i} | \vx, {y_i}} \Big] d\vz_{y_i} \\ 
    =& \sum_{i = 0}^{m} \KL(Q_{\vPhi}(\vz_{y_{i}} | \vx, y_{i})  \| P_{\vTheta}(\vz_{y_{i}} | y_i))
\end{align}

\subsection{E\hspace{10pt}Support Set Size}
\begin{table}[!ht]
\centering
\begin{tabular}{ccccccccccccc}
\hline

\hline
            & \multicolumn{4}{c}{Label   Generator} & \multicolumn{4}{c}{Conditional Text Generator} & \multicolumn{4}{c}{Classifier} \\\cmidrule(r){2-5}\cmidrule(r){6-9}\cmidrule(r){10-13}
Support Size & Jacc     & Acc      & Corr     & Comp    & Jacc     & Acc      & Corr     & Comp     & Jacc     & Acc     & Corr    & Comp    \\
\hline
0           & 54.40   & 23.56   & 41.78   & 44.59   & 58.83    & 31.41    & 47.74    & 49.31    & 58.83    & 32.37   & 49.96   & 48.39   \\
50          & 59.40    & 32.09    & 48.06    & 50.08   & 59.40    & 32.09    & 48.06    & 50.08    & 59.40    & 32.09   & 48.06   & 50.08   \\
100         & 58.26    & 30.02    & 47.32    & 48.40   & 59.02    & 31.70    & 48.50    & 49.06    & 60.67    & 33.31   & 50.64   & 50.77   \\
250         & 59.47    & 31.61    & 48.32    & 50.24   & 59.87    & 31.92    & 48.80    & 50.53    & 60.24    & 35.23   & 52.96   & 48.83  \\
\hline

\hline

\end{tabular}
\caption{The performance of the BERT classifier, based on four evaluation metrics, when different support data sizes are applied to three distinct modules.}
\label{samplesizefull}
\end{table}
The supplementary results on how the support set size affects the performance of the label generator, conditional text generator, and classifier in terms of all metrics are presented in Table~\ref{samplesizefull}.

\subsection{F\hspace{10pt}Synthetic Data Size}

\begin{figure}[!h]
    \centering
    \includegraphics[width=0.5\linewidth]{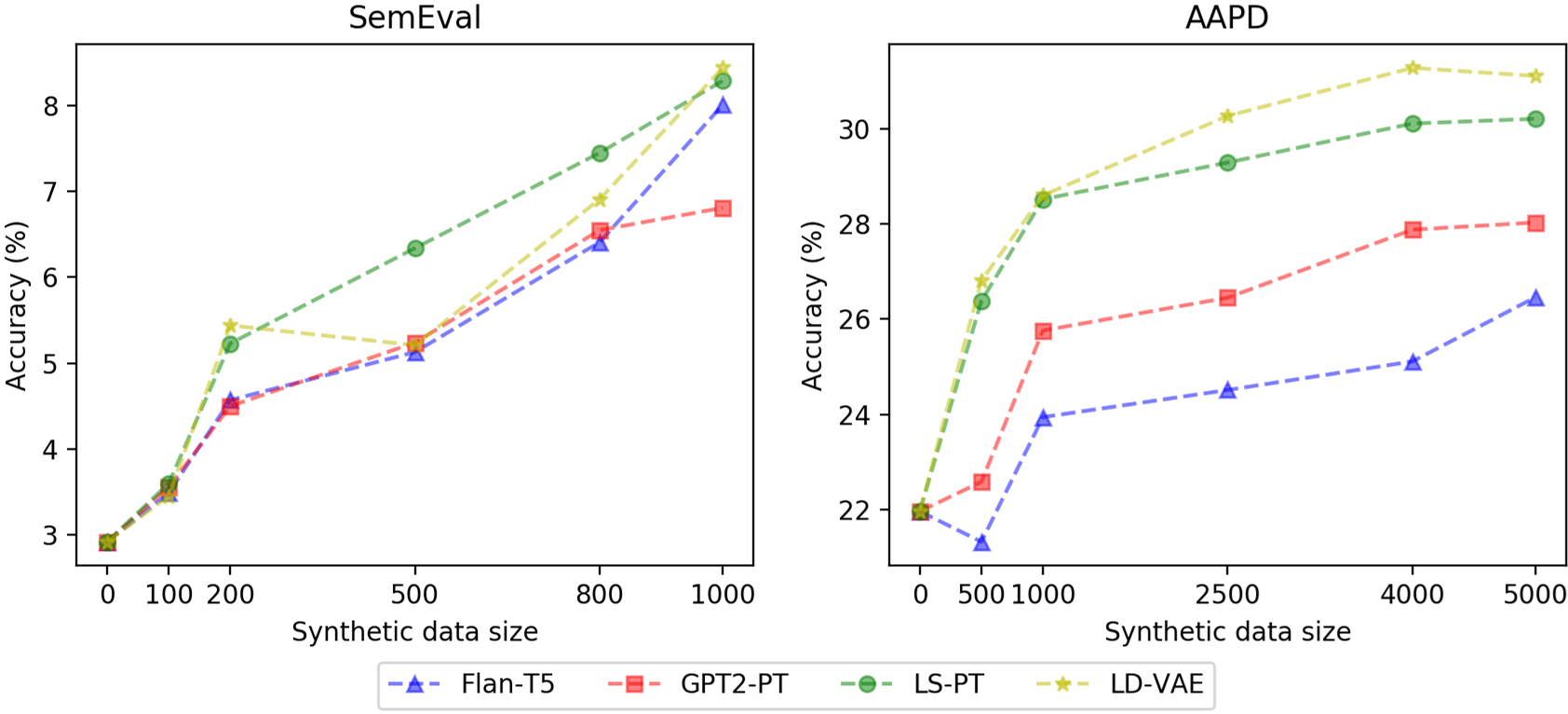}
    \caption{Accuracies of the BERT classifier using synthetic data of various sizes generated by different generators.
    }
    \label{datasize}
\end{figure}

In Figure~\ref{datasize}, classification accuracy improves as synthetic data size increases. Our LS-PT and LD-VAE models produce high-quality texts, evidenced by consistently higher accuracy. With just 1000 synthetic data points, our two models achieve 28\% accuracies on AAPD, while GPT2-PT needs 5000, highlighting the ability of our methods to generate high-quality instances with new label compositions.

\subsection{G\hspace{10pt}Disentanglement}
\begin{table}[!ht]
\centering
\label{VAE}
\begin{tabular}{llcccc}
\toprule[0.9pt]
\multicolumn{2}{c}{}                  & Jacc  & Acc   & Corr  & Comp  \\
\hline
\multicolumn{2}{l}{LD-VAE}   & 58.50 & 31.11 & 48.67 & 48.09 \\
\multicolumn{2}{c}{VAE} & 56.29 & 26.06 & 45.94 & 46.03 \\
\bottomrule[0.9pt]
\end{tabular}
\caption{The classification results of LD-VAE and vanilla VAE which using unconditional priors on AAPD.}
\label{vae}
\end{table}

\begin{figure}[!ht]
    \centering
    \includegraphics[width=0.45\linewidth]{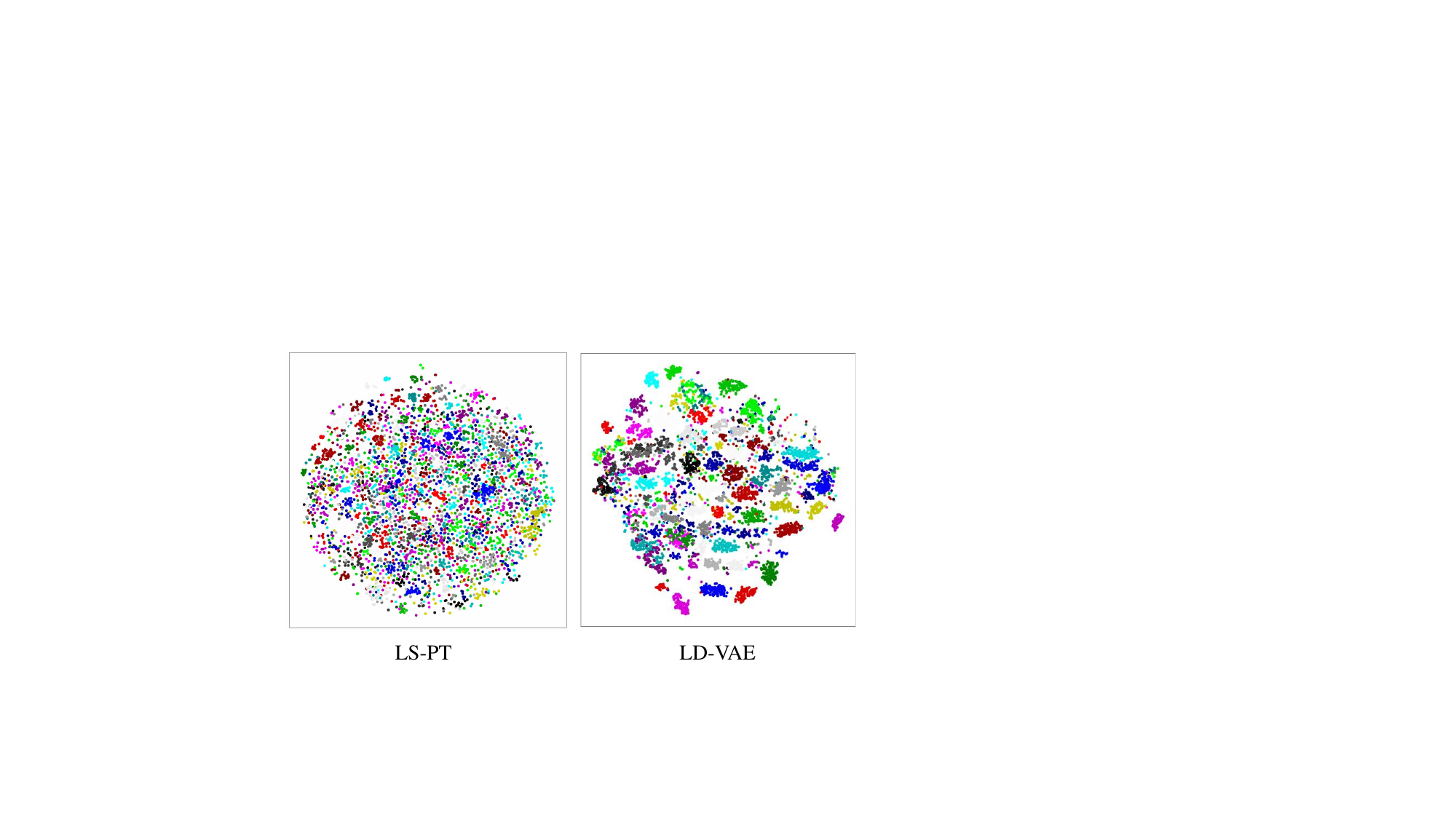}
    \caption{T-SNE visualization showing the label phrase representations encoded by GPT2 from both LS-PT and LD-VAE.
    }
    \label{disentangle_phrase}
\end{figure}


We replaced the conditional disentangled priors in LD-VAE with the unconditional priors found in the vanilla VAE. As illustrated in Table~\ref{vae}, a performance decline ranging from 3.9\% to 16.2\% across the four metrics shows the importance of our disentangled label priors.

In Figure~\ref{disentangle_phrase}, we depict the label phrase representations for LS-PT and LD-VAE encoded by GPT2, instead of the prefix vectors shown in Figure~\ref{disentangle}. The label phrase representations of both LS-PT and LD-VAE are more disentangled and clustered than those of GPT2-PT, even though all three are encoded with the pre-trained Transformers of GPT2. This suggests that our method enhances the disentanglement of Transformer-encoded representations. However, it is worth noting that the label phrase representations of LS-PT are not as well-clustered as those of Flan-T5, even though LS-PT still outperforms Flan-T5 when it comes to improving the MLTC classifier performance. We hypothesize that the superior performance of LS-PT primarily comes from its use of label-specific representations in GPT2's prefix vectors.

\subsection{H\hspace{10pt}Error Analysis}
\subsubsection{CG Classification Errors.}
Fine-tuning MLTC models often leads to overfitting the label composition distribution in training sets, resulting in the models learning spurious correlations~\cite{pmlr-v119-sagawa20a}. Take SemEval as an instance: the label composition ``joy, love, optimism'' appears over 600 times in the training set. Yet, the classifier correctly predicts test instances with the label ``joy, love'' only 4 out of 283 times, with most predictions erroneously being ``joy, love, optimism''. Such behaviour suggests that the classifier has learned these misleading label correlations, hampering CG in MLTC. This type of error is less prevalent in AAPD and IMDB compared to SemEval, possibly because emotion labels in SemEval exhibit a stronger correlation.

Moreover, MLTC models exhibit limited robustness to input perturbations, further affecting CG capability. Consider the challenges of emotion classification, such as the implicit emotional expression problem. Given the input ``Having a movie day with my favourite today — god I love my lil goth bean,'' the BERT classifier correctly identifies the unseen label combination ``joy, love''. But when presented with a semantically similar sentence ``Beautiful morning at the beach on Anna Maria Island with my wife,'' which implicitly conveys ``joy, love'', the classifier fails.

\subsubsection{Generation Errors}
\begin{table}[!th]
\centering
\resizebox{0.98\textwidth}{!}{
\begin{tabular}{lc}

\hline

\hline
Flan-T5 & \makecell[l]{\textbf{joy, sadness} : This is an absolute sadness.   \#pilgrimage \#SkyNews \\ \textbf{Comedy, Crime, Drama} : During a traffic stop, he and his   accomplice in the robbery of dozens of antiques,\\ including the rarest of the   1950s and the most expensive of them, are blown up. In the meantime, the two men \\ are looking for something to do with Freddie's memory.}                        \\
\hline
GPT2-PT & \makecell[l]{\textbf{joy, sadness} : @winklesdore @chrythesweet @vox\_babysitters   @hannypol@shanehansen It's good to be with \\ you and laugh. \#laugh \\ \textbf{Comedy,   Crime, Drama} : A former cop-turned-witnesses, a girl named Amanda, is a woman  of few words \\ and a mother of five, who is in a very bad way about her own   life. In order to save her life she decides to go \\ undercover and get out.} \\
\hline

GPT3.5   & \makecell[l]{\textbf{joy, sadness} : Just got some bad news... Feeling heartbroken right now. \#devastated \#sad\\
\textbf{Comedy, Crime, Drama} : A hilarious and thrilling ride, this comedy crime drama keeps you guessing until \\ the very end.With a missing daughter, stolen sculptures, and a treasure hunt, the plot is full of unexpected twists \\ and turns. The talented cast delivers stellar performances, making this a must-watch for fans of all genres.
}  \\
\hline
LS-PT   & \makecell[l]{\textbf{joy, sadness} : I'm still laughing but I've got so many regrets. \#lost \\
\textbf{Comedy, Crime, Drama} : After a violent robbery at the local bar the couple and the man are accused of murder. \\ The police find the pair and find it is a gang of two armed men who attacked them at gunpoint but they're not \\ found the first time.
}                                                                                                                       \\
\hline
LD-VAE  & \makecell[l]{\textbf{joy, sadness} : I'm sorry to hear about the loss of a friend, but it's worth it to know you're still alive. \\*
\textbf{Comedy, Crime, Drama} : A teenage boy is arrested for murder after a drunken party, and his former girlfriend \\ (played by Lisa McBride) finds out and moves into her home. The pair are joined by 'Rashad', a beautiful young \\ woman and a budding star (Sharon Hill) who is trying to find a way out of this horrible situation. However, in \\ reality the young man isn't helping anyone and RASHAD is caught in an impasse.\\*}     \\
\hline

\hline
\end{tabular}}
\caption{Synthetic examples generated by different generators.}
\label{examples}
\end{table}


Our data augmentation approach effectively mitigates the learning of spurious correlations and enhances model robustness. It achieves this by generating high-quality instances that authentically represent the target label compositions. For instance, as demonstrated in Table~\ref{examples}, our LD-VAE model can adeptly generate sentences such as ``joy, sadness: I'm sorry to hear about the loss of a friend, but it's comforting to know you're still here,'' which captures both ``joy'' and ``sadness'' emotions accurately.

Although our filtering mechanism excels at identifying and excluding instances with inaccurate labels, we still notice another common generation exists post-filtering. The generated text sometimes only reflects a subset of the provided label composition. This could potentially lower the performance of the classifiers. For example, Flan-T5 produced the instance ``\textit{joy, sadness: This is absolute sadness. \#pilgrimage \#SkyNews}'', wherein only the ``sadness'' emotion is evident. Similarly, GPT2-PT yielded a movie review, ``\textit{Comedy, Crime, Drama: A former cop-turned-witness named Amanda, a woman of few words and a mother of five, finds herself in dire straits. To save herself, she decides to go undercover.}'', which seems more aligned with just ``Crime'' and ``Drama'' genres.

\subsection{I\hspace{10pt}Human Evaluation}
\begin{table}[!ht]
\centering
\begin{tabular}{ccccccccc}
\toprule[0.9pt]
        &               & \multicolumn{2}{c}{SemEval}   &           &               & \multicolumn{2}{c}{IMDB} &           \\
        \cmidrule(r){2-5}\cmidrule(r){6-9}
        & Acc           & Corr          & Comp          & Avg. R. & Acc           & Corr            & Comp   & Avg. R. \\
        \hline
Flan-T5 & 34\%          & 78\%          & \textbf{37\%} & 2.67  & 37\%          & 88\%            & 38\%   & 4.33         \\
GPT3.5  & 28\%          & \textbf{91\%} & 29\%          & 3         & 39\%          & \textbf{92\%}   & 42\%   & 2.33         \\
GPT2-PT & 18\%          & 76\%          & 18\%          & 5         & 39\%          & 87\%            & 41\%   & 3.67  \\
LS-PT   & 36\%          & 83\%          & 35\%          & 2.33  & 40\%          & 85\%            & \textbf{49\%}   & 2.67  \\
LD-VAE  & \textbf{37\%} & 80\%          & 36\%          & \textbf{2}  & \textbf{41\%} & 90\%            & 47\%   & \textbf{1.67} \\
\bottomrule[0.9pt]
\end{tabular}

\centering
\caption{Human evaluation of Accuracy, Correctness, and Completeness across labels of utterances from various generators. 'Avg. R.' denotes the average rank in terms of different metrics for each generator on the test sets of each dataset.}
\label{human}
\end{table}

We conducted human evaluation by randomly sampling 50 examples and engaged three students to perform a human evaluation of the utterances generated by various generators. The results, as presented in Table~\ref{human}, show that LD-VAE and LS-PT achieved noticeably higher levels of accuracy and completeness compared to the other baseline methods. The average rankings across different metrics also indicate that LD-VAE produced the highest-quality utterances, while LS-PT generated the second-highest quality, both of which more comprehensively covered the semantics associated with the given label composition than other generators. Furthermore, the Correctness and Completeness scores reveal that all generators tend to generate utterances that convey the semantics of a subset of the ground truth labels rather than their superset.

\end{document}